%% file: main.tex
\documentclass{article} % For LaTeX2e
\usepackage{iclr2027_conference,times}

\input{math_commands.tex}

\usepackage{hyperref}
\usepackage{url}
\usepackage{xspace}

\usepackage{soul}
\newcommand{\method}{\textsc{Dwell}\xspace}

\usepackage{booktabs}       % professional-quality tables
\usepackage{amsfonts}       % blackboard math symbols
\usepackage{nicefrac}       % compact symbols for 1/2, etc.
\usepackage{microtype}      % microtypography
\usepackage{xcolor}  % colors

\usepackage{amssymb}      % extra math symbols, e.g., \mathbb, \triangleq-related symbols
\usepackage{mathtools}    % extensions/fixes for amsmath
\usepackage{bm}           % bold math symbols, e.g., \bm{v}

\usepackage{algorithm}    % floating algorithm environment
\usepackage{algpseudocode} % algorithmic commands: \State, \For, \If, \Return, etc.

\usepackage{graphicx}     % include figures
\usepackage{caption}      % caption customization
\usepackage{subcaption}   % subfigures, if needed

\usepackage{array}        % better table column control
\usepackage{multirow}     % multirow table cells
\usepackage{enumitem}     % compact and customized lists
\usepackage{wrapfig}

\usepackage{amsmath}

\usepackage[most]{tcolorbox} % 核心宏包，[most] 会加载绝大多数常用库

\usepackage{dsfont}

\usepackage{subcaption}

\usepackage{algpseudocode}

\usepackage{makecell}
\usepackage{listings}
\usepackage[table]{xcolor} 

\usepackage{adjustbox}

\usepackage{colortbl}
\usepackage{tabularx}
\usepackage{siunitx} 
\usepackage{xspace}  

\usepackage{algpseudocode}
\algnewcommand{\LineComment}[1]{\State $\triangleright$ #1}

\title{Vision-Language Agents Can Learn in Their Own Latent Worlds}

\author{Haoxi Li$^{1}$,
Sikai Bai$^{1}$,
Jianfei Ma$^{2}$,
Qinglin Hou$^{3}$,
Xiaocheng Lu$^{1}$,
Tao Han$^{1}$,\\
\textbf{Jie Zhang$^{1,\dagger}$},
\textbf{Song Guo$^{1,\dagger}$} \\[0.6em]
{\normalfont
$^{1}$Department of Computer Science and Engineering, HKUST, Hong Kong, China} \\
{\normalfont
$^{2}$School of Computing, National University of Singapore,  Singapore} \\
{\normalfont
$^{3}$Department of Computer Science, University of Southern California, Los Angeles, USA} \\[0.4em]
{\normalfont
$^{\dagger}$Corresponding authors}
}

\iclrfinalcopy

\begin{document}

\maketitle

\begin{abstract}
% new

Learning from experience is essential for building the next generation of agents that continually expand their capabilities. 
Pretrained vision-language models (VLMs) already internalize rich world knowledge, yet this knowledge is rarely leveraged to drive policy improvement through online interaction. 
Realizing this potential efficiently remains challenging, as policy updates shift the agent’s interaction distribution, requiring the world model to adapt accordingly, while continually updating large pretrained models is computationally expensive.
To address this challenge, we introduce \method, a model-based reinforcement learning (RL) framework that reframes a VLM as a recurrent latent transition operator, endowing it with the ability to predict, adjust, and act in latent space. Leveraging the VLM’s own representations and its dual role as both world model and policy, the agent continually learns from real interactions to anticipate and evaluate action outcomes, and from latent imagination to improve its behavior.
Across five interactive visual environments, \method achieves the strongest overall performance among trainable methods while substantially reducing the compute required for imagination, showing that effective world models for VLM agents need not reconstruct the future---they need to predict futures that are useful for action. Our code is publicly available at \url{https://github.com/Tom111-lhx/Dwell}.

\end{abstract}

\section{Introduction}

AI agents~\citep{openai2026gpt6astra,team2026kimi} have acquired broad capabilities through large-scale pretraining on human-generated data~\citep{brown2020language,hoffmann2022training} and post-training with human demonstrations and preference feedback~\citep{ziegler2019fine,ouyang2022training}.
Extending this progress beyond the limits of human-provided data~\citep{villalobos2024position} motivates agents that continually learn from experience generated through their own interactions with the environment~\citep{silver2025welcome}.
Recent work~\citep{wang2026vagen,li2026you} has taken steps in this direction by using reinforcement learning (RL) to improve agents' multi-step decision-making with environmental feedback.
However, the substantial environment interaction often required by model-free approaches can make experience collection a bottleneck to scaling~\citep{janner2019trust,chandra2025diwa}. This raises a central question: \emph{how can agents achieve greater policy improvement from limited real experience?}

World models offer a way to learn more from real experience by predicting action outcomes and generating imagined trajectories for policy learning.
For vision-language model (VLM) agents, models that synthesize observations have already served as training environments across embodied and web tasks~\citep{zhu2026wmpo,fang2025webevolver}. In contrast, imagination in a compact latent space avoids reconstructing observations at every predicted step~\citep{hafner2019dream,hafner2023mastering}, while predictive representation learning can emphasize predictable structure over incidental visual detail~\citep{assran2023self,maes2026leworldmodel}.
Pretrained VLMs already encode rich world knowledge, suggesting that their own representations could support latent imagination for policy learning.

Accurate prediction alone does not guarantee compatibility: a world model's latent states may not be readily usable by a VLM policy, even when they faithfully capture action outcomes~\citep{assran2025v,chen2026vl}.
A further difficulty arises as policy updates shift the agent's interaction distribution toward states and actions poorly covered by the world model's training data~\citep{janner2019trust}.
In these regions, errors in predicted transitions and rewards can bias subsequent policy updates, so gains in imagination may not translate into better behavior in the environment~\citep{liu2026world}.
Adapting to this evolving distribution can also be computationally expensive when it requires repeated updates to a large pretrained VLM.
Together, these challenges motivate a shared representational basis for prediction and decision-making, coupled with efficient, continual correction of transition and reward predictions from real interaction.

We introduce \method, a model-based RL framework that enables VLMs to predict, adjust, and act in their own latent space. By recasting the VLM as a recurrent latent transition operator, we equip it to reason about the state of the world from interaction history and improve its behavior through latent imagination. 
Following the joint-embedding principle~\citep{assran2023self}, our method draws on the VLM's rich world knowledge to actively infer a state representation in an abstract and compact space, rather than focusing on every detail of the observation.
Exploiting the VLM's dual role as a world model and policy, we instantiate both components using separate lightweight LoRA adapters~\citep{hu2021lora} on the same VLM backbone, allowing independent, parameter-efficient adaptation while sharing a common representation interface. Together, recurrence and duality provide a unified framework for modeling and acting in dynamic environments through online learning.

We conduct comprehensive experiments across five interactive visual environments spanning spatial planning, embodied navigation, manipulation,
and long-horizon household tasks. under the reported training protocol, \textsc{Dwell} achieves the superior performance among trainable
methods, while substantially reducing the computational cost of imagination relative to a world-model baseline that reconstructs observations.
In summary, the main contributions of the paper are as follows:
\begin{itemize}[leftmargin=*]
    \item \textbf{Learning and acting in a shared latent world.}
    We recast a pretrained VLM as a recurrent latent transition operator, unifying world modeling and policy learning through a common belief interface.

    \item \textbf{Online policy improvement through latent imagination.}
    We develop a learning loop in which real interactions continually refine the world and reward models, while imagined trajectories guide policy improvement.
    
    \item \textbf{Empirical gains in performance and efficiency.} Our proposed method achieves strong overall performance among the evaluated trainable baselines under the reported training settings, while reducing imagination cost relative to a world model that synthesizes observations.
    
\end{itemize}

\section{Method}
\label{sec:method}

A self-evolving agent should anticipate the consequences of its actions and use feedback from interaction to continually refine its predictions and behavior. 
Although VLMs have acquired rich world knowledge, they tend to be brittle when exposed to dynamic environments or when learning from experience. 
Our method, \method, anchors a VLM as a recurrent latent transition operator, morphing a static perceiver into an active predictor. Given streams of observations and actions from online interaction, it infers the most explainable states in the VLM's own latent space, which can in turn be used for model-based planning in the absence of feedback. This cycle of interaction, refinement, and improvement enables the agent to continuously expand its understanding of the environment while adapting its actions accordingly.

\paragraph{Interaction Setting.}
We model this online interaction as a partially observable Markov decision process (POMDP)~\cite{aastrom1965optimal, kaelbling1998planning}, defined by a tuple $(\mathcal{S}, \mathcal{A}, \mathcal{O}, T, R, Z, \gamma)$, where $\mathcal{S}$, $\mathcal{A}$, and $\mathcal{O}$ correspond to the state, action, and observation spaces. At each step $t$, the agent takes an action $a_{t} \sim \pi_{\theta}(a_{t} | b_{t})$ based on its belief $b_{t}$, a sufficient statistic computed from the history $h_t=(o_{0},a_{0},\dots,o_{t-1},a_{t-1},o_{t})$. The environment then transitions to a hidden state $s_{t+1} \sim T(s_{t+1} \mid s_t,a_t)$ and emits a visual observation $o_{t+1} \sim Z(o_{t+1} \mid s_{t+1},a_t)$ and reward $r_{t + 1} = R(s_t,a_t,s_{t+1})$. Since the state is unobserved, the VLM agent must form a belief over it and learns a policy $\pi_\theta$ to maximize the expected cumulative discounted return:
\begin{equation}
    \mathcal{J}(\theta) = \mathbb{E}_{\pi_\theta, T, Z} \left[ \sum_{t=0}^{\infty} \gamma^t R(s_t,a_t,s_{t+1}) \right],
    \label{eq:rl_loss}
\end{equation}
where $\gamma \in [0, 1)$ is the discount factor. The belief $b_t$ upon which the agent acts is updated deterministically as $b_{t} = \tau(b_{t - 1}, a_{t - 1}, o_{t})$ using a belief update function $\tau$ given an initial belief $b_{0}$. Depending on the model specification, the belief can take either a probabilistic or deterministic form.

\subsection{Recurrence and Duality: A Common Interface}
\label{sec:belief}

\paragraph{Recurrent Latent Model.} Learning from experience requires connecting two sources of state information: what the agent observes and what it can anticipate from its
actions. 
Let $\nu(o_t)$ denote the ordered sequence of visual tokens extracted by the VLM from observation $o_t$, and $\mathbf z_t\in\mathbb R^{K\times D}$ represents the agent's latent state, comprising $K$ slots of dimension $D$.
% Let $\nu(o_t)$ denote the VLM's ordered visual tokens and $\mathbf z_t\in\mathbb R^{K\times D}$ its state consisting of $K$ latent slots. 
Given the previous state $\mathbf z_{t-1}$ and action $a_{t-1}$, we define two recurrent updates:
% observation-conditioned state inference and action-conditioned prediction as:
\begin{equation}
\label{eq:state}
\begin{aligned}
\mathbf z_t
  &=\Sigma_\phi\bigl(\mathbf z_{t-1},a_{t-1},\nu(o_t)\bigr),\\
\widehat{\mathbf z}_{t}
  &=\Psi_\phi\bigl(\mathbf z_{t-1},a_{t-1}\bigr).
\end{aligned}
\end{equation}
The observation-conditioned update $\Sigma_\phi$ incorporates the realized outcome of $a_{t-1}$ into the agent's evolving belief, whereas the action-conditioned predictor $\Psi_\phi$ anticipates that outcome without access to $o_t$. 
We refer to their outputs as the \emph{posterior} state $\mathbf z_t$ and the \emph{prior} state $\widehat{\mathbf z}_t$, respectively.
Here, these two terms distinguish the information available to each update: both states are deterministic latent representations rather than probability distributions.

Conceptually, if we follow a joint embedding predictive architecture (JEPA)~\citep{assran2023self}, our formulation generalizes it by allowing the incorporation of historical information. This can be seen by removing $\mathbf{z}_{t-1}$ and $a_{t-1}$ from the posterior and setting $\Sigma_\phi = I$. When the observation alone is not a sufficient statistic, our formulation induces better representations while improving the separability of the latent states by anchoring their representations to the observation. Compared with reconstruction-based methods, ours operates in the latent space without focusing on every detail of the observation, thereby improving computational efficiency.

\paragraph{VLM as World Model and Policy.} 

We instantiate the world model and policy using separate LoRA adapters~\citep{hu2021lora} on a shared, frozen pretrained VLM backbone.
Within the world model, posterior inference $\Sigma_\phi$ and prior prediction $\Psi_\phi$ share the same trainable parameters $\phi$, whereas the policy $\pi_\theta$ uses a distinct set of adapter parameters $\theta$.
This separation supports independent, parameter-efficient adaptation of prediction and decision-making while keeping the pretrained backbone fixed.

The latent slots retain their ordering across updates, providing a common interface for transition prediction, reward evaluation, and action selection. The policy $\pi_\theta$ conditions on posterior states during real interaction and on predicted states during latent imagination.
Thus, predicted states can be passed directly to the next state transition and decision without decoding them into the raw observation and re-encoding them. Initialization and parameterization are detailed in Appendix~\ref{app:state_training}.

\subsection{Dynamics, Representation, and Reward Learning}
\label{sec:prediction}

\begin{figure}[t]
    \centering
    \includegraphics[width=1.0\textwidth]{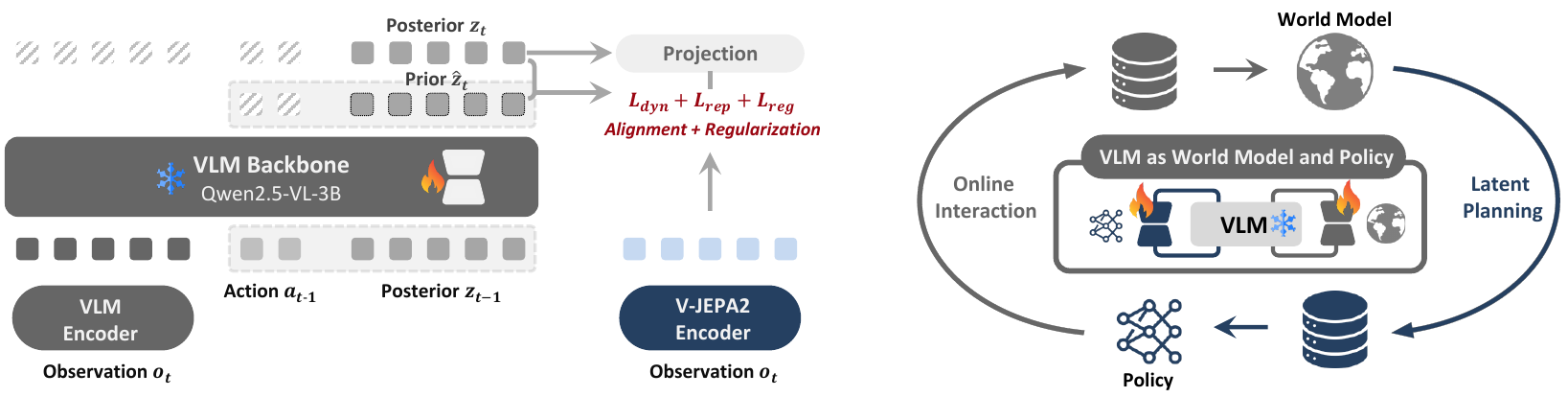}
    \caption{
    \textbf{Overview of \method.}
    \textbf{Left:} The VLM learns a recurrent latent world by matching action-conditioned priors to observation-conditioned posteriors, while grounding the latent states with a frozen V-JEPA2 encoder.
    \textbf{Right:} The same VLM backbone serves as both world model and policy, alternating online interaction with latent planning for policy improvement.
}
    \label{fig:main}
\end{figure}

Learning a world model improves sample efficiency, enables planning and foresight, and enhances adaptability to environmental changes. The central questions are: \textbf{(1)} what state representation a good world model requires; and \textbf{(2)} how to make reliable predictions into the future to guide action selection. When faced with complex or high-dimensional observations, a naive approach is to optimize for reconstruction fidelity in the original space. However, this can be problematic, as it focuses too much on information that is redundant or irrelevant to solving the task, while also being computationally inefficient. VLMs can internalize rich knowledge in their latent space, providing us with a good starting point to build upon. We can leverage this internalized knowledge by operating in a compact latent space while attending only to the information most relevant to decision-making.

\paragraph{Dynamics Learning.}
When explicit feedback is absent, the world model needs to make reliable predictions into the future for planning and foresight. The posterior, in contrast, infers the states that best explain the current observation given the history, providing a natural learning target for the prior and thereby equipping the VLM with the ability to predict and adjust.

Given sequences $\tau = (a_t, o_{t+1}, a_{t+1}, o_{t+2}, \dots, o_{t + T})$ from a replay buffer $\mathcal D$, we learn the prior by:
\begin{equation}
\label{eq:dyn}
\mathcal L_{\mathrm{dyn}}(\phi)
=
\mathbb E_{\tau \sim \mathcal D}\!\left[\sum\limits_{k=1}^{T}\left\|\widehat{\mathbf z}_{t+k} -\operatorname{sg}(\mathbf z_{t+k})\right\|_{\mathrm{F}}^2\right],
\end{equation}
where $\operatorname{sg}$ denotes stop-gradient.

\paragraph{Representation Learning.} However, agreement between the posterior and prior alone does not establish their connection to observations; they can freely minimize the error while capturing no useful information for decision-making. While we can anchor the posterior to the VLM's visual encoder, a representation that is well-suited for supervised learning may not readily transfer to online learning. Therefore, we turn to frozen pretrained visual encoders $\mathbf e_t=\mathrm{E}(o_t)$ such as V-JEPA2 through a learned slot-wise projection $\Gamma_\phi$ for posterior learning:
\begin{equation}
\label{eq:representation}
\mathcal L_{\mathrm{rep}}(\phi)
=
\mathbb E_{\tau \sim \mathcal D}\!\left[\sum\limits_{k=1}^{T}\mathbf{d}\!\left(\Gamma_\phi(\mathbf z_{t + k}),\operatorname{sg}(\mathbf e_{t + k})\right)\right],
\end{equation}
where, for $\mathbf d$, we use the averaged cosine distance over corresponding slots. Notably, despite the projection, learning remains in a compact representation space rather than the raw observation space, akin to JEPA, thereby avoiding expensive reconstruction.

\paragraph{Reward Learning.} Although it is straightforward to learn the reward from the posterior, planning in the latent space with the prior presents a gap, which widens as the planning horizon increases due to compounding errors. For this reason, we propose to learn the reward from a pure prior lookahead given the future actions $a_{t:t+L}$, starting from a posterior $\mathbf{z}_t$. We denote this compositional mapping as $\Pi(\mathbf{z}_t, a_{t:t+L}; L)$, which results in a trailing prior $\widehat{\mathbf{z}}_{t+L+1}$. This prior combines two ends: the history encoded in the posterior and the future actions, from which we predict the reward $\widehat{r}_{t+L+1} = R_{\omega}(\widehat{\mathbf{z}}_{t+L+1})$. Over the trajectory, we sample a lookahead horizon up to the maximum length $T$ and apply $\Pi$ at $t_k := t + \lfloor k - L - 1 \rfloor$ for a given trailing step $t+k$:
\begin{equation}
\label{eq:reward}
\mathcal L_{\mathrm{rew}}(\omega)
= \mathbb E_{\tau \sim \mathcal D, L \sim [T]}\!\left[\sum\limits_{k=1}^{T}\! \ell_r\left(R_{\omega}(\Pi(\mathbf{z}_{t_k}, a_{t_k:t_k+L}; L)), r_{t+k}\right)\right],
\end{equation}
where $\ell_r$ is the mean squared error for continuous rewards and the cross-entropy loss for discrete ones. This approach balances long-horizon reward prediction in the latent space with the use of historical information, providing a spectrum of stable learning for model-based planning. Notably, when $L = 0$, it reduces to per-step prediction, which accounts for only one action at a time; when $L \geq T$, the prior is decoupled from the posterior and makes predictions on its own, ignoring any historical information.

Putting it all together, with an auxiliary regularizer~\citep{maes2026leworldmodel} to discourage representation collapse, the objective for learning the world model is:
\begin{equation}
\label{eq:wm}
\mathcal J_{\mathrm{WM}}
= \mathcal L_{\mathrm{dyn}}
+\alpha \mathcal L_{\mathrm{rep}}
+\beta \mathcal L_{\mathrm{rew}}
+\eta \mathcal L_{\mathrm{reg}},
\end{equation}
where $\alpha, \beta,$ and $\eta$ indicate the relative strength of each term.

\subsection{Model-based Latent Planning.}
\label{sec:online}
A world model is a mirror of reality, allowing us to simulate without constant interaction with the environment. Planning with a world model requires anticipating future rewards to identify better actions, which can be compactly represented by a value function. However, the interleaved nature of the world model and value function necessitates a reliable initial value estimate to be backpropagated during planning, especially at the beginning of training.

\paragraph{Bridge the Gap.}

Learning to predict a state does not yet establish how much future reward it yields. With finite imagination, early policy updates also depend on value estimates that have not yet been fitted to task returns. We therefore warm up the critic on real experience using the $n$-step return $G_t^{(n)}$ before allowing it to guide policy improvement, while keeping the actor and world model fixed.

\paragraph{Planning with World Model.}
Starting from a posterior state $\widehat{\mathbf z}_0=\mathbf z_t$ obtained by unrolling Eq.~\ref{eq:state}, we generate a batch $\mathcal{B}$ of imagined experiences $\{a_{h - 1}, \widehat{\mathbf{z}}_{h}, \widehat{r}_{h}\}_{h=1}^{H}$ of length $H$ using the prior and the current policy $\pi_{\theta}$:
\begin{equation}
\label{eq:imagine}
a_{h-1} \sim \pi_{\theta}(\cdot\mid\widehat{\mathbf z}_{h-1}),\quad \widehat{\mathbf z}_{h} =\Psi_\phi(\widehat{\mathbf z}_{h-1},a_{h-1}),\quad \widehat r_h=R_\omega(\widehat{\mathbf z}_{h}), \quad \forall h \in [H].
\end{equation}

Given the imagined batch $\mathcal{B}$, in principle, we can use any actor-critic algorithm to learn both the value function and the policy. To improve sample efficiency and reduce variance, we adopt PPO~\citep{schulman2017proximal} to update both.

\paragraph{Close the Loop: Iterative Online Learning.}
The learned policy returns to the real environment, reflecting the world model's predictions, to collect new experiences and revise its beliefs. This cycle of alternating between interaction and learning, which subsumes the interplay between the world model and the policy and is centered around the VLM, enables continual adaptation and learning from experience at the boundary between the known and the unknown, the history and the future. Through this process, the agent improves its behavior iteratively through trial and error, as summarized in Algorithm~\ref{alg:dwell}.

\section{Experiments}
\label{sec:experiments}

We evaluate \method across diverse interactive visual environments to examine whether latent imagination improves task performance and learning efficiency while remaining computationally practical.

\subsection{Experimental Setup}
\label{sec:experimental_setup}

\paragraph{Environments.}
We evaluate on five interactive visual environments spanning complementary decision-making regimes:
\textsc{Sokoban}, \textsc{FrozenLake}, \textsc{Navigation}, \textsc{PrimitiveSkill}, and \textsc{EB-ALFRED}.
\textsc{Sokoban}~\citep{SchraderSokoban2018} and \textsc{FrozenLake}~\citep{towers2024gymnasium} test spatial reasoning and multi-step planning in discrete visual environments;
\textsc{Navigation}~\citep{yang2025embodiedbench,kolve2017ai2} evaluates embodied navigation from first-person observations;
\textsc{PrimitiveSkill}~\citep{tao2024maniskill3,nasiriany2022augmenting} tests visually grounded manipulation;
and \textsc{EB-ALFRED}~\citep{yang2025embodiedbench} extends the evaluation to long-horizon compositional embodied tasks.
Together, these environments cover increasingly complex forms of partial observability, action consequence prediction, and long-horizon decision making.
We use task success rate as the primary metric and, unless otherwise specified, report the mean and standard deviation over three independent training seeds.

\paragraph{Baselines.}
We use \texttt{Qwen2.5-VL-3B}~\citep{Qwen-VL} as the unified VLM backbone for all trainable agent methods. We compare \method against model-free online RL with \textsc{PPO}~\citep{schulman2017proximal} and \textsc{GRPO}~\citep{guo2025deepseek}, which optimize the policy directly from environmental feedback without imagined experience. For model-based RL, we adapt \textsc{LeWorldModel} to our online agent setting by connecting its learned latent dynamics to the same VLM policy and online interaction pipeline; the adaptation is detailed in Appendix~\ref{app:baseline_details}. We additionally report strong off-the-shelf VLMs as reference points without online adaptation.

\paragraph{Implementation Details.}
We keep the VLM backbone frozen and equip the policy and world model with separate trainable LoRA adapters.
We use a frozen V-JEPA2~\citep{assran2025v} encoder as the default visual target for representation grounding. We use LoRA rank $r=32$ with scaling factor $\alpha=64$. Full hyperparameters and environment-specific configurations are provided in Appendix~\ref{app:env-settings}.

\subsection{Overall Performance}
\begin{table*}[t]
    \centering
    \caption{
    Overall performance across five interactive visual environments. All trainable methods use \texttt{Qwen2.5-VL-3B} as the VLM backbone and are trained under the same environmental feedback and interaction budget.
    Off-the-shelf VLMs are evaluated without online adaptation.
    Bold indicates the best performance among trainable methods.
    }
    \begin{adjustbox}{width=\linewidth}
    \setlength\tabcolsep{1pt}
    \setlength\extrarowheight{3pt}
    \arrayrulecolor[gray]{0.7}

    \begin{tabular}{l|c|c|cc|c|cccc|c|c|c}
        \toprule
        \multirow{2}{*}{\textbf{Model/Method}}
        & \multirow{2}{*}{\textbf{Sokoban}}
        & \multirow{2}{*}{\textbf{FrozenLake}}
        & \multicolumn{3}{c|}{\textbf{Navigation}}
        & \multicolumn{5}{c|}{\textbf{PrimitiveSkill}}
        & \multirow{2}{*}{\textbf{EB-ALFRED}}
        & \multirow{2}{*}{\textbf{Overall}} \\
        \cmidrule(lr){4-11}
        & & &
        Base & Common & \textbf{Average}
        & Place & Stack & Drawer & Align & \textbf{Average}
        & & \\
        \hline

        % =========================================================
        % Off-the-shelf VLMs
        % =========================================================
        \multicolumn{13}{c}{Open-Source VLMs} \\
        \hline

        \color{gray}Qwen2.5-VL-72B~\citep{Qwen-VL}
        & \color{gray}0.20
        & \color{gray}0.44
        & \color{gray}0.70
        & \color{gray}0.77
        & \color{gray}0.74
        & \color{gray}\textbf{1.00}
        & \color{gray}0.50
        & \color{gray}0.00
        & \color{gray}\textbf{1.00}
        & \color{gray}0.63
        & \color{gray}0.39
        & \color{gray} 0.48\\

        \color{gray}Qwen2.5-VL-7B~\citep{Qwen-VL}
        & \color{gray}0.14
        & \color{gray}0.14
        & \color{gray}0.33
        & \color{gray}0.38
        & \color{gray}0.35
        & \color{gray}0.00
        & \color{gray}0.00
        & \color{gray}0.00
        & \color{gray}0.75
        & \color{gray}0.19
        & \color{gray}0.47
        & \color{gray} 0.26\\

        Qwen2.5-VL-3B~\citep{Qwen-VL}
        & 0.13
        & 0.14
        & 0.20
        & 0.26
        & 0.23
        & 0.00
        & 0.00
        & 0.00
        & 0.00
        & 0.00
        & 0.01
        & 0.10 \\

        \color{gray}VLM-R1-3B \citep{shen2025vlm}
        & \color{gray}0.16
        & \color{gray}0.15
        & \color{gray}0.33
        & \color{gray}0.34
        & \color{gray}0.34
        & \color{gray}0.00
        & \color{gray}0.00
        & \color{gray}0.00
        & \color{gray}0.00
        & \color{gray}0.00
        & \color{gray}0.02
        & \color{gray}0.13 \\
        
        \hline
        \multicolumn{13}{c}{Proprietary VLMs} \\
        \hline

        \color{gray}GPT-4o~\citep{hurst2024gpt}
        & \color{gray}0.43
        & \color{gray}0.54
        & \color{gray}0.75
        & \color{gray}0.69
        & \color{gray}0.72
        & \color{gray}0.50
        & \color{gray}0.63
        & \color{gray}0.00
        & \color{gray}0.88
        & \color{gray}0.50
        & \color{gray}0.56
        & \color{gray}0.55 \\

        \color{gray}Claude 4.5 Sonnet \cite{Claude4_5}
        & \color{gray}0.31
        & \color{gray}0.80
        & \color{gray}0.67
        & \color{gray}0.67
        & \color{gray}0.67
        & \color{gray}0.63
        & \color{gray}0.50
        & \color{gray}0.00
        & \color{gray}\textbf{1.00}
        & \color{gray}0.53
        & \color{gray}\textbf{0.71}
        & \color{gray}0.60 \\

        \hline

        % =========================================================
        % Model-Free Online RL
        % =========================================================
        \rowcolor{gray!25}
        \multicolumn{13}{c}{
        Model-Free Online RL (Backbone: Qwen2.5-VL-3B)
        } \\

        \rowcolor{gray!10}
        PPO~\citep{schulman2017proximal}
        & $0.29_{\scriptscriptstyle\pm0.125}$
        & $0.22_{\scriptscriptstyle\pm0.036}$
        & $0.28_{\scriptscriptstyle\pm0.072}$ %0.28 0.072(0.32,0.2,0.33)
        & $0.27_{\scriptscriptstyle\pm0.026}$%0.27 0.026(0.25 0.26 0.34)
        & $0.28$ 
        & $0$
        & $0$
        & $0$
        & $0.29_{\scriptscriptstyle\pm0.508}$%0.29 0.508
        & $0.07$ 
        & $0.06_{\scriptscriptstyle\pm0.034}$%0.47,0.17,0.2
        & 0.18 \\

        \rowcolor{gray!10}
        GRPO~\citep{guo2025deepseek}
        & $0.21_{\scriptscriptstyle\pm0.040}$%0.21 0.040 (0.25,0.17,0.21)
        & $0.19_{\scriptscriptstyle\pm0.052}$%0.19 0.052 (0.23,0.15,0.20) 
        & $0.74_{\scriptscriptstyle\pm0.100}$%0.74 0.100 (0.83,0.75,0.63)
        & $0.71_{\scriptscriptstyle\pm0.053}$%0.71 0.053 (0.70,0.75,0.69)
        & $0.73$%0.73 0.068 (0.765,0.75,0.6) 
        & $0$
        & $0$
        & $0$
        & $0.67_{\scriptscriptstyle\pm0.578}$%0.67 0.578
        & $0.17$
        & $0.10_{\scriptscriptstyle\pm0.026}$
        & 0.28 \\

        \hline

        % =========================================================
        % Model-Based RL
        % =========================================================
        \rowcolor{gray!25}
        \multicolumn{13}{c}{
        Model-Based RL (Backbone: Qwen2.5-VL-3B)
        } \\

        \rowcolor{gray!10}
        LeWorldModel~\citep{maes2026leworldmodel}
        & $0.44_{\scriptscriptstyle\pm0.021}$%0.43,0.46,0.42
        & $0.29_{\scriptscriptstyle\pm0.061}$%0.28,0.36,0.24
        & $0.30_{\scriptscriptstyle\pm0.037}$%0.32,0.26,0.33
        & $0.28_{\scriptscriptstyle\pm0.0493}$
        & $0.29$
        & $0$
        & $0.17_{\scriptscriptstyle\pm0.289}$%0,0,0.5
        & $0$
        & $0$
        & $0.04$
        & $0.13_{\scriptscriptstyle\pm0.030}$
        &  0.24\\

        \rowcolor{gray!10}
        \textsc{\method}
        & \textbf{$0.53_{\scriptscriptstyle\pm0.012}$}%0.52,0.54,0.54
        & \textbf{$0.51_{\scriptscriptstyle\pm0.021}$}%0.49, 0.52,0.53
        & \textbf{$0.37_{\scriptscriptstyle\pm0.036}$}%0.38,
        & \textbf{$0.40_{\scriptscriptstyle\pm0.0208}$}
        & \textbf{$0.39$}
        & \textbf{$0.33_{\scriptscriptstyle\pm0.288}$}
        & \textbf{$0.21_{\scriptscriptstyle\pm0.259}$}%0,0.5,0.13
        & \textbf{$0$}
        & \textbf{$0.29_{\scriptscriptstyle\pm0.508}$}
        & \textbf{$0.28$}
        & \textbf{$0.18_{\scriptscriptstyle\pm0.048}$}
        & \textbf{0.38} \\

        \bottomrule
    \end{tabular}
    \end{adjustbox}
    \label{tab:main_results}
\end{table*}

\paragraph{Results.}
As shown in Table~\ref{tab:main_results}, \method achieves the best overall performance among trainable methods, substantially improving the same \texttt{Qwen2.5-VL-3B} initialization. The gains are most evident on \textsc{Sokoban}, \textsc{FrozenLake}, and the long-horizon \textsc{EB-ALFRED} tasks, where successful decisions depend strongly on anticipating how current actions alter future states. This pattern supports our central hypothesis that action-conditioned latent imagination can turn limited real experience into useful additional training experience.

The benefit is task dependent rather than uniform. Model-free GRPO remains stronger on \textsc{Navigation} and slightly better on \textsc{PrimitiveSkill}. We attribute this difference to two distinct regimes: in navigation, immediate visual observations often provide strong action cues, reducing the marginal benefit of explicit future-state prediction; in contact-rich manipulation, useful imagination requires finer-grained dynamics that are harder to preserve in a compact latent rollout. In contrast, \method consistently improves over the adapted \textsc{LeWorldModel} baseline, suggesting that effective model-based learning depends not only on learning latent dynamics, but also on making predicted states directly usable by the VLM policy. We examine the efficiency of these gains and the role of each
component below.

\subsection{Learning and Computational Efficiency}
\label{sec:efficiency}

\begin{figure*}[htbp]
    \centering
    \includegraphics[width=\textwidth]{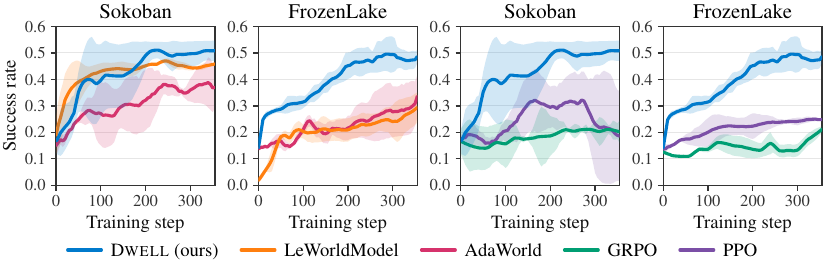}
    \caption{
    \textbf{Learning Efficiency.}
    Success rate throughout training on \textsc{Sokoban} and
    \textsc{FrozenLake}.
    \textit{Left}: comparison with model-based approaches.
    \textit{Right}: comparison with model-free approaches.
    }
    \label{fig:learning_efficiency}
\end{figure*}

We next study whether the gains of \method can be obtained efficiently, both in terms of learning from online interaction and the computation required to support imagination.
Figure~\ref{fig:learning_efficiency} reports learning curves under the same training protocol.
For the model-based comparison, we additionally include \textsc{AdaWorld}~\citep{gao2025adaworld}, an adaptable generative world model based on latent actions, as a representative visual-generation-based world model.
We adapt its official implementation to our evaluation setting; implementation details are provided in Appendix~\ref{app:baseline_details}.

\paragraph{Learning Efficiency.}
Across both \textsc{Sokoban} and \textsc{FrozenLake}, \method reaches higher success rates than the alternative model-based approaches.
The advantage is particularly clear on \textsc{FrozenLake}, where the competing methods improve only gradually while \method continues to benefit from interaction. These results suggest that recurrent latent imagination provides useful additional learning signal from experience already collected, rather than relying solely on new environment rollouts.

\begin{figure*}[htbp]
    \centering
    \includegraphics[width=\textwidth]{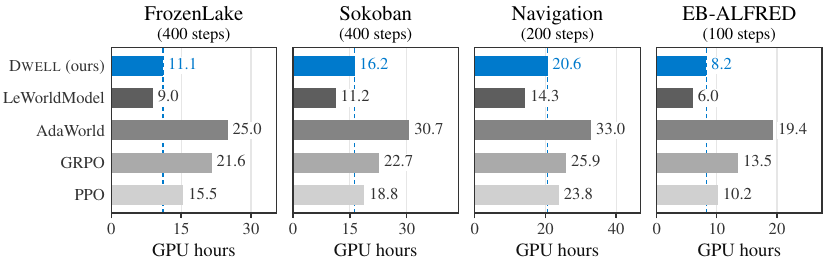}
    \caption{
    \textbf{Computational Efficiency.}
    End-to-end GPU-hours required to reach the reported training step on
    \textsc{FrozenLake}, \textsc{Sokoban}, \textsc{Navigation}, and
    \textsc{EB-ALFRED}.
    GPU-hours include all computation incurred during online training,
    including world-model updates, imagined rollouts, and policy/value optimization.
    All methods are evaluated under the same environment-specific training schedule and hardware configuration.
    }
    \label{fig:compute_efficiency}
\end{figure*}

\paragraph{Computational Efficiency.}
We further measure the end-to-end GPU-hours required to reach the final training step used in each environment, including all GPU computation incurred by online training, such as world-model updates, imagined rollouts, and policy/value optimization.
As shown in Figure~\ref{fig:compute_efficiency}, \textsc{LeWorldModel} remains the least expensive approach due to its compact latent dynamics.
Nevertheless, \method consistently requires substantially less computation than \textsc{AdaWorld}, reducing GPU-hours by approximately $38$--$58\%$ across the four evaluated environments.
It also incurs lower total compute than PPO and GRPO under the same training schedule.
Taken together with the performance results in Table~\ref{tab:main_results}, these findings indicate that \method achieves a favorable performance--compute trade-off: it retains the efficiency of latent-space imagination while avoiding the considerably higher cost of generating future visual observations.

\subsection{Ablation Studies}
\label{sec:ablation}

We ablate two aspects of \method that are central to learning from latent
imagination: the supervision used to construct predictive latent states and
the continual adaptation of the world model during online policy learning.

\begin{figure*}[htbp]
    \centering
    \includegraphics[width=0.96\textwidth]{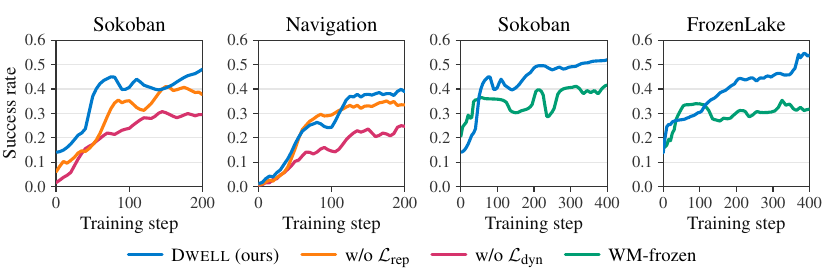}
    \caption{
    \textbf{Ablation Studies.}
    Left: effect of removing the representation-grounding objective
    $\mathcal{L}_{\mathrm{rep}}$ or the predictive dynamics objective
    $\mathcal{L}_{\mathrm{dyn}}$.
    Right: effect of freezing the world model after initialization instead
    of continually updating it with newly collected experience.
    Predictive dynamics provide the primary supervision for useful latent
    imagination, while representation grounding and continual world-model
    adaptation further improve policy learning.
    }
    \label{fig:ablation}
\end{figure*}

\paragraph{Role of the World Model Objectives.}
As shown in Figure~\ref{fig:ablation}, removing either predictive dynamics learning or representation grounding
degrades performance on both \textsc{Sokoban} and \textsc{Navigation}.
The degradation is substantially larger without
$\mathcal{L}_{\mathrm{dyn}}$, indicating that direct supervision of
action-conditioned future states is essential for producing useful imagined
trajectories.
Removing $\mathcal{L}_{\mathrm{rep}}$ yields a smaller but consistent drop,
showing that predictive consistency alone is insufficient and benefits from
anchoring the latent state to perceptually meaningful visual representations.
Together, the two objectives play complementary roles:
$\mathcal{L}_{\mathrm{dyn}}$ learns how the state evolves under actions,
whereas $\mathcal{L}_{\mathrm{rep}}$ constrains what information the state
should preserve.

\paragraph{Effect of Online World Model Adaptation.}
We further freeze the world model after its initial fitting and continue
policy optimization with the resulting fixed dynamics.
The frozen model remains competitive early in training, but its performance
quickly saturates on both \textsc{Sokoban} and \textsc{FrozenLake}, whereas
the full \method continues to improve as additional interaction is collected.
This behavior is consistent with a policy-induced distribution shift:
as the policy changes, it visits state--action regions that are increasingly
different from those used to fit the initial model.
Continually refreshing the world model with newly collected experience
therefore keeps imagined trajectories aligned with the evolving policy
distribution and enables sustained policy improvement.

\section{Related Work}

\paragraph{Latent World Models for Behavior Learning.}
Model-based RL has long used learned dynamics to augment real experience with
prediction and planning. Dreamer learns policies from trajectories imagined in
a compact latent state space, while TD-MPC and its successors optimize behavior
with decoder-free latent world models~\citep{hafner2019dream,
hafner2023mastering,hansen2024td}.
More recent predictive-representation approaches replace observation
reconstruction with latent prediction, enabling action-conditioned planning or
policy learning directly in representation space
~\citep{assran2025v,bagatella2026td}.
These methods establish that control need not operate on reconstructed
observations. Our focus is complementary: we study how such a latent learning
interface can be built into a VLM agent, so that the representations
used for real decisions also form the state space of imagined policy learning.

\paragraph{World Models for Agent Learning.}
Generative world models have recently become practical training environments
for large-model agents. In embodied settings, video world models synthesize
action-conditioned futures for VLA policy optimization
~\citep{zhu2026wmpo,jia2026dreamplan}, while learned web simulators generate
interaction experience for self-improvement in browser agents
~\citep{fang2025webevolver,ding2026dynaweb}.
A common design is to adapt the simulator to policy behavior and
then hold it fixed while optimizing the policy.
This separation makes model-generated experience reusable, but the simulator
cannot incorporate new evidence arising during the subsequent policy updates.
World-VLA-Loop~\citep{liu2026world} takes a step toward closing this loop by iteratively adding new
policy rollouts to world-model training; however, policy
learning still proceeds through autoregressively generated visual observations and inherits their long-horizon drift.
\method instead uses the VLM's predicted beliefs themselves as the
interaction states: policy learning requires neither future-image generation
nor re-encoding, and the latent world is refreshed from newly collected real
experience throughout training.

\paragraph{World Modeling within VLM Agents.}
A parallel line of work develops world-modeling capabilities inside the agent
itself. VLM representations can provide semantically informed state features
for control, and explicit state estimation and transition reasoning can improve
multi-turn visual decision-making~\citep{chen2024vision,wang2026vagen}.
Other approaches incorporate future prediction more directly into action
learning:
VLA-JEPA uses latent state prediction to learn action-relevant
representations, and unified world--action models jointly predict observations
and actions~\citep{sun2026vla,cen2025worldvla}.
In these methods, world modeling primarily shapes the representation, reasoning
process, or action predictor.
\method gives the predicted state a different operational role: it is
the state on which the policy continues to act and learn.
A single frozen VLM backbone therefore supports two trainable roles through
separate adapters: one learns action-conditioned latent dynamics, while the
other learns the policy, and real interaction continually supervises both the
predicted dynamics and their task outcomes.

\section{Limitation and Discussion}

\paragraph{When Does Latent Imagination Help?}
The benefit of latent imagination is not uniform across the evaluated tasks.
While \method shows clear gains on \textsc{Sokoban}, \textsc{FrozenLake}, and
\textsc{EB-ALFRED}, its advantage is smaller on \textsc{Navigation} and
\textsc{PrimitiveSkill}. This pattern is consistent with the hypothesis that
imagination is most useful when predicted future states provide information
beyond what is already available from the current observation, whereas its
benefit may diminish when action selection is largely observation-driven or
depends on dynamics that are difficult to capture compactly. A natural next
step is therefore to make imagination selective, allowing the agent to decide
when and how far to imagine rather than applying a fixed rollout horizon
uniformly across states and tasks.

\paragraph{What Must a Latent World Preserve?}
Operating directly in a compact latent space avoids the cost of reconstructing
future observations, but introduces a fundamental trade-off between abstraction
and decision sufficiency. Representations that discard visually incidental
details may also remove fine-grained dynamics that later become relevant for
control, particularly in contact-rich interaction. Moreover, errors in latent
transitions and reward prediction can accumulate during planning,
making long-horizon rollouts increasingly unreliable. Our current deterministic
latent dynamics do not explicitly represent such uncertainty, leaving the
policy unable to distinguish a confident prediction from an ambiguous one.
These limitations suggest that stronger latent world models may need to learn
not only what is predictable, but what must be preserved for downstream
decisions, together with uncertainty estimates that determine how far imagined
experience should be trusted.

\paragraph{Scope and Scalability.}
Our current evidence is limited to a single VLM backbone and five simulated
interactive visual environments, and therefore does not yet establish how the
same design scales with model capacity, richer action spaces, or real-world
interaction. In addition, although the latent world is represented within the
VLM, its representation is grounded using a frozen pretrained visual encoder;
whether equally useful predictive structure can emerge solely from interaction
remains an open question. Latent imagination is also not computationally free:
while \method substantially reduces end-to-end GPU-hours relative to a
generative world model, simpler latent dynamics remain cheaper, revealing a
performance--compute trade-off rather than an absolute efficiency advantage.
Extending this framework to larger VLMs, transferable world representations,
and persistent learning across tasks and environments would be important steps
toward agents that continually refine a shared model of their experience.

\section{Conclusion}

We presented \method, a model-based RL framework that enables VLM agents to learn from experience in their own latent worlds. Rather than treating the world model as a separate simulator that reconstructs future observations, \method recasts the VLM as a recurrent latent transition operator and uses a shared representational interface for prediction, value estimation, and action. By interleaving real interaction with latent imagination, the agent can reuse limited environmental experience to improve its policy without generating future observations explicitly. Across diverse interactive visual environments, \method achieves strong overall performance under controlled interaction budgets while substantially reducing the computational cost of imagination relative to generative world models. Taken together, these results suggest that a useful world model need not reproduce the world itself; it needs to predict a future that the agent can directly act and learn upon.

\bibliography{iclr2027_conference}
\bibliographystyle{iclr2027_conference}

\newpage
\appendix
\section{Appendix}

\providecommand{\dwellcheck}[2]{}

\subsection{Algorithm}
  \begin{algorithm}[H]
  \caption{\method}
  \label{alg:dwell}
  \begin{algorithmic}[1]
  \Require posterior $\Sigma_\phi$ and prior $\Psi_\phi$; reward model $R_\omega$; actor $\pi_\theta$; critic $V_\psi$; planning horizon $H$; world model updates $N$; latent imaginations $L$; PPO epochs $E$
  \State Prefill $\mathcal{D}$ with a random policy
  \State Pretrain world model using Eq.~\ref{eq:wm}
  \State Warm up $V_\psi$ using $n$-step return $G_t^{(n)}$ as in Eq.~\ref{eq:real_targets}
  \While{not converged}
      \LineComment{\textbf{Online Interaction}}
      \For{time step $t = 1, \dots, T$}
          \State Observe $o_t$ and infer posterior state $\Sigma_\phi\bigl(\mathbf z_{t-1},a_{t-1},\nu(o_t)\bigr)$
          \State Execute $\pi_\theta(a_t | \mathbf{z}_t)$ in the environment, receive $o_{t+1}$, reward $r_{t + 1}$, and termination $d_{t + 1}$
          \State Tokenize the transition $(o_t, a_t, o_{t+1}, r_{t+1}, d_{t+1})$ and append it to the replay buffer $\mathcal D$
      \EndFor
      \vspace{0.5em}
      \LineComment{\textbf{World Model Learning}}
      \For{iteration $n=1, 2, \dots,N$}
          \State Compute loss function as in Eq.~\ref{eq:wm}
          \[
          \mathcal J_{\rm WM}
          =
          \mathcal L_{\rm dyn}
          + \alpha\mathcal L_{\rm rep}
          + \beta\mathcal L_{\rm rew}
          + \eta\mathcal L_{\rm reg}.
          \]
          \State Update world model $(\phi, \omega)$ with $\nabla_{\phi,\omega}\mathcal J_{\rm WM}(\phi,\omega)$
      \EndFor
      \vspace{0.5em}
      \LineComment{\textbf{Model-based Latent Planning}}
      \State Freeze $(\phi, \omega)$, snapshot $\theta_{\rm old}\gets\theta$, and initialize imagined batch $\mathcal B\gets\emptyset$
      \For{imagination $l=1,\ldots,L$}
          \State Initialize a batch of $B$ posteriors $\mathbf{z}_0$ from $\tau \sim \mathcal{D}, |\tau| > H$ via $t \sim \mathrm{Unif}\{0, \dots, |\tau|-H-1\}$ posterior steps
          \State Unroll the prior model for $H$ steps over the batch following Eq.~\ref{eq:imagine} and store them in $\mathcal B$
      \EndFor
      \vspace{0.5em}
      \LineComment{\textbf{Actor-Critic Learning}}
      \State Compute advantages $\widehat A_h$ and $\lambda$-returns $\widehat G_h^\lambda$ for $\mathcal B$ using Eq.~\ref{eq:gae}
      \For{epoch $e=1,\dots, E$}
          \State Shuffle $\mathcal B$ into minibatches
          \For{each minibatch}
              \State Jointly update actor and critic via $\nabla_{\theta, \psi}\mathcal{J}(\theta, \psi)$ following Eq.~\ref{eq:ppo}
          \EndFor
      \EndFor
  \EndWhile
  \end{algorithmic}
\end{algorithm}

\subsection{Baseline}
\label{app:baseline_details}

All trainable methods use the same \texttt{Qwen2.5-VL-3B} initialization, environment
wrappers, action vocabulary, train/evaluation splits, real-interaction budget,
and evaluation seeds.  PPO and GRPO optimize the environment reward directly.
For \textsc{LeWorldModel}, we retain its latent-dynamics objective but adapt its
predicted latent state to the same actor--critic interface used by \textsc{Dwell}
through a learned state mapping.  This mapping only reconciles representation
shape and action indexing; it does not expose observations, rewards, or labels
that are unavailable to the other methods.  Before policy optimization,
\textsc{LeWorldModel} uses the same actor/critic initialization stages as
\textsc{Dwell}: behavior cloning is applied only in environments for which it
is listed in Table~\ref{tab:hyperparams}, and the critic is warmed up on the
same real-replay return targets.  Reward scales, action mappings, termination
rules, PPO clipping and GAE conventions, update budgets, and evaluation
intervals are matched environment by environment.  The methods differ only in
the learned world-model representation and the mechanism used to produce
imagined transitions.  Hyperparameters are selected without access to the
evaluation levels, and no evaluation trajectory is used for training, replay,
warm-up, or model selection.  \textsc{AdaWorld} is adapted under the same
interaction, optimization, and evaluation budgets while retaining its native
generative world-model objective.

\subsection{Additional Ablation Study Results}

\paragraph{Shared latent interface.}
A central design choice in \method is that the world model and policy operate
on the same latent state interface: predicted states can therefore be consumed
directly by the policy without an additional representation conversion.
To isolate the effect of this design, we construct a \emph{bridged} variant
in which the world model retains its own latent representation and a
lightweight learned adapter maps predicted states into a policy-specific
latent space before action selection.
All other components, including the VLM backbone, world-model objectives,
policy optimization, imagination horizon, and online training schedule, are
kept unchanged.

As shown in Table~\ref{tab:shared_interface}, separating the prediction and
decision spaces consistently degrades performance across all three evaluated
environments. The drop is most pronounced on \textsc{Sokoban}, while similar
degradations are observed on \textsc{FrozenLake} and \textsc{Navigation}.
These results provide evidence that the benefit of \method does not arise
solely from learning latent dynamics: making imagined states directly
compatible with the policy's decision representation also facilitates
learning from imagined experience.

\begin{table}[t]
    \centering
    \caption{
    \textbf{Effect of the shared latent interface.}
    We compare the default \method, where predicted latent states are consumed
    directly by the policy, with a bridged variant that maps world-model states
    into a separate policy latent space through a learned adapter.
    Results are success rates averaged over three training seeds.
    }
    \label{tab:shared_interface}
    \small
    \setlength{\tabcolsep}{5pt}
    \renewcommand{\arraystretch}{1.15}
    \begin{tabular}{lccc}
        \toprule
        \textbf{Variant}
        & \textbf{Sokoban}
        & \textbf{FrozenLake}
        & \textbf{Navigation} \\
        \midrule
        \method\ (Shared)
        & \textbf{$0.53_{\scriptscriptstyle\pm0.012}$}
        & \textbf{$0.51_{\scriptscriptstyle\pm0.021}$}
        & \textbf{$0.39$} \\
        
        Bridged Latent Interface
        & $0.38_{\scriptscriptstyle\pm0.138}$
        & $0.42_{\scriptscriptstyle\pm0.097}$
        & $0.31_{\scriptscriptstyle\pm0.017}$ \\
        \bottomrule
    \end{tabular}
\end{table}

% ===================================================================
% Environment settings, reward assignment, and key hyperparameters.
% ===================================================================
\providecommand{\tbd}[1]{\textcolor{red}{[#1]}}

\subsection{Environment Settings}
\label{app:env-settings}

Table~\ref{tab:environment_settings} reports one environment decision per
step.  A PrimitiveSkill language turn may contain at most two sequential
primitive commands; we therefore report both turns and the corresponding
maximum number of executed primitives.  PrimitiveSkill comprises four tasks
(Place, Stack, Drawer, and Align), with 16 held-out episodes per task.

\begin{table}[t]
  \centering
  \caption{Environment settings. Horizon is the maximum number of training /
  evaluation decisions. PrimitiveSkill observations also include task-state text;
  each episode has three language turns with at most two primitive commands per
  turn (at most six in total), and its 64 evaluation episodes cover four tasks
  (16 each).}
  \label{tab:environment_settings}
  \small
  \begin{tabular}{llllr}
    \toprule
    Environment & Observation & Actions & Horizon & Eval. episodes \\
    \midrule
    Sokoban & top-down RGB, $6\!\times\!6$, one box & 4 moves & 25/25 & 256 \\
    FrozenLake & top-down RGB, $4\!\times\!4$ & 4 moves & 25/25 & 128 \\
    Navigation & first-person RGB, instruction & 8 motions & 30/10 & 120 \\
    PrimitiveSkill & RGB, instruction & 3 parameterized skills & 3/3 turns & 64 \\
    EB-ALFRED & first-person RGB, instruction & 162 skills & 30/30 & 20 \\
    \bottomrule
  \end{tabular}
\end{table}

\subsection{Reward assignment}
\label{app:reward_assignment}

Rewards enter training in two ways.  Model-free baselines optimize the real
task reward directly.  \textsc{Dwell} does not insert the real environment
reward into imagined trajectories: real transitions supervise the feedback
head, and imagined rewards are read from that head on predicted latent states.
The calibrated threshold or floor is selected on a training-only calibration
split and frozen before evaluation.  For EB-ALFRED, the final ranked reward
checkpoint gives
\[
  c = 0.3357132077,
  \qquad
  \widetilde p = 0.1p + 0.9\,\operatorname{clip}
  \left(\frac{p-c}{1-c},0,1\right).
\]
For Sokoban, the calibrated termination threshold is
$\tau=0.4968768060$ (reported as $0.497$ in the main text).

\begin{table}[t]
  \centering
  \caption{Reward assignment. $p$ is the predicted success probability, $\tau$ a
  calibrated threshold, $\widetilde p$ a damped version of $p$, and $t$ the
  imagination step. For PrimitiveSkill, the step reward is a $0.1$ format reward
  plus $\Delta s$, the increase in the highest completed task stage, and the
  feedback head also predicts the stage.}
  \label{tab:reward_assignment}
  \small
  \begin{tabular}{lcclll}
    \toprule
    & \multicolumn{2}{c}{Real reward} & \multicolumn{3}{c}{\textsc{Dwell} in imagination} \\
    \cmidrule(lr){2-3}\cmidrule(l){4-6}
    Environment & Success & Step & Label & Reward & Episode end \\
    \midrule
    Sokoban & $+10$ & $-0.1$ & binary & $1.1\,\mathbf 1[p\ge\tau]-0.01$ & $p\ge\tau$ \\
    FrozenLake & $+10$ & $0$ & 3-class & $0.1\,\mathbf 1[p_{\mathrm{goal}}\ge0.5]$ & goal or hole \\
    Navigation & $+10$ & $0$ & binary & $1.1\,p-0.1$ & cutoff at $H$ \\
    PrimitiveSkill & $+1$ & $0.1+\Delta s$ & binary, stage & $1.1\,p-0.1$ & success or $H$ \\
    EB-ALFRED & $+1$ & $-0.1$ & binary & $1.1\,\widetilde p\,\mathbf 1[t=H]-0.1$ & cutoff at $H$ \\
    \bottomrule
  \end{tabular}
\end{table}

\paragraph{Offline replay sampling.}
For every environment, the offline episode set is fixed before optimization,
uses training-only seeds, and is disjoint from all evaluation levels and
seeds.  Episodes are sampled uniformly at random without replacement when the
fixed offline subset is formed, and replay episodes and valid within-episode
windows are sampled uniformly during training; no reward-, success-, or
difficulty-prioritized sampling is used.  Collection policy and replay
sampling are distinct: Sokoban and FrozenLake use random collection, whereas
Navigation uses goal-directed, noisy goal-directed, and random collection in a
$3{:}2{:}1$ ratio and EB-ALFRED uses scripted, noisy-scripted, and random
collection in a $4{:}1{:}1$ ratio.

\paragraph{PrimitiveSkill protocol.}
The recommended complete PrimitiveSkill setting is: 20,000 offline episodes
(5,000 per task), random collection, 5,000 world-model steps with batch size
4 and learning rate $10^{-4}$, truncated-BPTT window 3, open-loop horizon 3
with weight 0.25, $\lambda_{ep}/\lambda_{var}=0.05/0.005$, an MLP(128)
feedback head trained on horizons 1--3 with batch size 256 and learning rate
$10^{-4}$, no behavior-cloning warm-up, critic learning rate $10^{-4}$ with
40/1,000 minimum/maximum warm-up updates, sampled imagination with
$64\times1$ start-state/sequence batches and horizon 3, two PPO epochs with
minibatch size 16, actor learning rate $10^{-6}$, clip/target-KL
$0.2/0.02$, entropy weight 0.01, discount/GAE $0.99/0.95$, world-model
refresh every 20 actor updates using batch size 1 and learning rate $10^{-6}$,
four new real episodes every four updates, replay caps 20,000/10,000,
200 accepted actor updates, evaluation every 20 updates, and one NVIDIA H20
(96 GB) per run.

\subsection{Key Hyperparameters}
\label{app:hyperparams}

Table~\ref{tab:hyperparams} lists the settings that define \method's training
pipeline in each environment. Three protocol details matter for reading the results.
(i) Imagined actions are sampled from the current policy throughout latent
rollouts.
(ii) In \textsc{Sokoban} and \textsc{FrozenLake}, the training-step axis
counts accepted actor updates: a candidate update is rolled back
(actor, critic, and optimizer state) if action-ranking accuracy on held-out
real transitions falls below the acceptance threshold.
(iii) \textsc{EB-ALFRED} initializes the actor by behavior cloning on
successful scripted demonstrations and retains a behavior-cloning term
during PPO.

\begin{table}[h]
\centering
\caption{Key hyperparameters of \method.}
\label{tab:hyperparams}
\small
\renewcommand{\arraystretch}{1.2}
\setlength{\tabcolsep}{4pt}
\begin{tabular}{@{}lcccc@{}}
\toprule
Setting & Sokoban & FrozenLake & Navigation & EB-ALFRED \\
\midrule
\multicolumn{5}{@{}l}{\emph{Backbone and latent state}} \\
VLM backbone & \multicolumn{4}{c}{\texttt{Qwen2.5-VL-3B-Instruct} (frozen)} \\
LoRA rank / $\alpha$ / dropout & \multicolumn{4}{c}{$32$ / $64$ / $0.05$} \\
Latent slots $K$ / width $D$ & \multicolumn{4}{c}{$36$ / $2048$} \\
Visual teacher $\mathrm{E}$ & \multicolumn{4}{c}{\texttt{vjepa2-vitg-fpc64-384-ssv2} (frozen)} \\
\midrule
\multicolumn{5}{@{}l}{\emph{World-model fitting}} \\
Offline episodes & $5{,}000$ & $5{,}000$ & $5{,}000$ & $300$ \\
Collection policy & mixed & mixed & mixed & mixed \\
Steps / batch & \multicolumn{2}{c}{$2{,}000$ / $4$} & \multicolumn{2}{c}{$5{,}000$ / $4$} \\
Learning rate & \multicolumn{2}{c}{$10^{-5}$} & \multicolumn{2}{c}{$10^{-4}$} \\
Truncated-BPTT window & \multicolumn{2}{c}{$2$} & \multicolumn{2}{c}{$8$} \\
Open-loop horizon / weight & \multicolumn{2}{c}{--} & \multicolumn{2}{c}{$4$ / $0.25$} \\
Prior / posterior / delta weight & \multicolumn{4}{c}{$0.5$ / $0.1$ / $0.25$} \\
Regularizer $\mathcal L_{\mathrm{reg}}$ weight & \multicolumn{4}{c}{1} \\
\midrule
\multicolumn{5}{@{}l}{\emph{Feedback head}} \\
Architecture & linear & linear & MLP ($128$) & linear \\
Logged-action horizons $H_r$ & \multicolumn{3}{c}{$1$--$8$} & $1$--$4$ \\
Batch / learning rate & \multicolumn{4}{c}{$256$ / $10^{-4}$} \\
\midrule
\multicolumn{5}{@{}l}{\emph{Actor and critic initialization}} \\
Behavior-cloning steps & \multicolumn{2}{c}{$0$} & $0$ & $500$ \\
Critic learning rate & \multicolumn{2}{c}{$3\times10^{-5}$} & \multicolumn{2}{c}{$10^{-4}$} \\
Critic updates (min / max) & \multicolumn{2}{c}{$20$ / $100$} & $50$ / $200$ & $200$ / $1{,}000$ \\
\midrule
\multicolumn{5}{@{}l}{\emph{Imagined PPO}} \\
Imagination horizon $H$ & \multicolumn{3}{c}{$4$} & $15$ \\
Start states $\times$ sequences & $32\times16$ & $32\times16$ & $8\times64$ & $64\times1$ \\
Epochs / minibatch & \multicolumn{2}{c}{$2$ / $64$} & \multicolumn{2}{c}{$2$ / $16$} \\
Actor learning rate & $2\times10^{-6}$ & $2\times10^{-6}$ & $10^{-6}$ & $2\times10^{-6}$ \\
Critic learning rate & \multicolumn{2}{c}{$3\times10^{-5}$} & \multicolumn{2}{c}{$10^{-4}$} \\
Clip $\epsilon$ / target KL & \multicolumn{2}{c}{$0.1$ / $0.01$} & $0.1$ / $0.02$ & $0.2$ / $0.02$ \\
Entropy weight $\eta$ & \multicolumn{2}{c}{$0.005$} & \multicolumn{2}{c}{$0.01$} \\
Behavior-cloning weight & \multicolumn{3}{c}{$0$} & $0.1$ \\
Discount $\gamma$ / GAE $\lambda$ & \multicolumn{4}{c}{$0.99$ / $0.95$} \\
Acceptance: top-1 / pairwise / tolerance & \multicolumn{2}{c}{$0.60$ / $0.60$ / $0.05$} & $0.60$ / $0.60$ / $0.05$ & $0.60$ / $0.60$ / $0.05$ \\
\midrule
\multicolumn{5}{@{}l}{\emph{Online schedule}} \\
World model during PPO & \multicolumn{2}{c}{refreshed every $50$} & refreshed every $20$ & frozen \\
Refresh batch / learning rate & \multicolumn{2}{c}{$16$ / $10^{-6}$} & $1$ / $10^{-6}$ & -- \\
New real episodes & \multicolumn{2}{c}{$1$ per update} & $4$ per $4$ updates & none \\
Replay: offline / online cap & \multicolumn{2}{c}{$800$ / $800$} & $800$ / $800$  & -- \\
Accepted actor updates & 400 & 400 & 200 & 100 \\
Evaluation interval & \multicolumn{3}{c}{$10$} & $20$ \\
\midrule
GPU / GPUs per run & \multicolumn{4}{c}{ H20 / 8} \\
\bottomrule
\end{tabular}
\end{table}

Offline data for Navigation mixes goal-directed, noisy goal-directed, and random
episodes; for EB-ALFRED, scripted, noisy scripted, and random episodes in the
ratio $4{:}1{:}1$.

\subsection{World Model Learning}
\label{app:state_training}

\subsubsection{Regularization}
To prevent representational collapse, we apply an auxiliary regularizer
$\mathcal L_{\Diamond}$ to all valid posterior representations. Following SIGReg-style
regularization~\citep{balestriero2025lejepa}, we encourage random one-dimensional
projections of the latent distribution to match a standard Gaussian using the
Epps--Pulley statistic, together with a VICReg-style variance floor~\citep{bardes2021vicreg}. Specifically,
\begin{equation}
\mathcal L_{\mathrm{reg}}
=
\lambda_{\mathrm{ep}}
\frac{1}{M}\sum_{m=1}^{M}
\int
e^{-s^2}
\left|
\hat{\varphi}_{m}(s)-e^{-s^2/2}
\right|^2
\,\mathrm{d}s
+
\lambda_{\mathrm{var}}
\frac{1}{D}\sum_{d=1}^{D}
\max\!\left(0,\,1-\mathrm{std}_{d}\right),
\label{eq:latent_regularization}
\end{equation}
where $\hat{\varphi}_{m}$ denotes the empirical characteristic function
along the $m$-th random projection, and $\mathrm{std}_{d}$ is the empirical
standard deviation of the $d$-th latent dimension over valid posterior slots.
We use $\lambda_{\mathrm{ep}}=0.05$ and
$\lambda_{\mathrm{var}}=0.005$.

\subsection{Reward and Termination Learning}
\label{app:feedback}
Both the reward model and the terminal predictor are lightweight MLPs that take the latent state as input, using mean pooling to aggregate the latent slots. The output layer of the reward model depends on the type of task reward: for continuous rewards, we model the reward using a Gaussian distribution with constant variance, whereas for sparse rewards or rewards that can be represented as discrete events, we instead use a categorical distribution. The terminal predictor outputs a Bernoulli distribution. Once the termination is predicted, we define the continuation as $c_h = 1-\widehat{d}_h$.

\dwellcheck{AUTHOR-TASKS}{The audits bind only Sokoban and FrozenLake to local
code/configuration artifacts. Add navigation, EB-ALFRED, and other evaluated
tasks' actual label/readout/terminal rules and instruction inputs before
claiming this appendix covers all experiments. Do not extrapolate the
action spaces or classifier types of these two tasks.}

\subsection{Actor-Critic Learning}

\subsubsection{Critic Warm-up}
\label{app:critic_init}
Consider a real trajectory segment $\tau = \{a_{t - 1}, o_{t}, r_{t}, d_{t}\}_{t=1}^T$, we infer states $\mathbf z_{1:T}$ from the posterior and compute $n$-step return $G_t^{(n)}$ as:
\begin{equation}
\label{eq:real_targets}
\begin{aligned}
G_T^{(n)}&=V_{\psi}(\mathbf z_T),\\
G_{t}^{(n)}&=r_{t + 1}+\gamma(1-d_{t + 1})G_{t+1}^{(n)},
\quad t=T-1,\ldots,0,
\end{aligned}
\end{equation}
which are used as targets for critic warm-up to provide a good starting point for policy improvement.

\subsubsection{Model-based Latent Planning}
\label{app:ppo}
Although any actor-critic algorithm can be used for model-based planning, we adopt PPO~\citep{schulman2017proximal} to train both the actor and critic, as it allows the same batch to be reused multiple times, thereby improving sample efficiency.

\paragraph{Critic Learning}
Given a batch of imagined trajectories $\mathcal{B}$, we use generalized advantage estimation~\citep{schulman2015high} to estimate both advantage and $\lambda$-return simultaneously, reducing variance in both policy gradient and TD targets for stable learning. Set $\widehat A_H=0$ and compute backwards over $h=H-1,\ldots,0$:
\begin{equation}
\label{eq:gae}
\begin{aligned}
\delta_h
&=\widehat r_{h + 1}+\gamma c_{h + 1}V_{\psi}(\widehat{\mathbf z}_{h+1})
-V_{\psi}(\widehat{\mathbf z}_h),\\
\widehat A_h
&=\delta_h+\gamma\lambda c_{h + 1}\widehat A_{h+1},\\
G_h^\lambda
&=\widehat A_h+V_{\psi}(\widehat{\mathbf z}_h).
\end{aligned}
\end{equation}
We then fit the critic to these targets to capture long-term rewards alongside the policy gradient.

\paragraph{Actor Learning}
We snapshot the actor $\pi_{\theta}$ as the sampling policy for PPO and denote it as $\pi_{\mathrm{old}}$, from which we can define the importance ratio:
\begin{equation}
\label{eq:ratio}
\rho_h(\theta)
= \frac{\pi_\theta(a_h\mid\widehat{\mathbf z}_h)}{\pi_{\mathrm{old}}(a_h\mid\widehat{\mathbf z}_h)}.
\end{equation}
In summary, we have the overall objective for policy improvement:
\begin{equation}
\label{eq:ppo}
\begin{aligned}
\mathcal J(\theta,\psi)
=\mathbb E_{\mathcal B}\Big[
&-\min\!\left(
\rho_h(\theta)\widehat A_h,
\operatorname{clip}(\rho_h(\theta),1-\epsilon,1+\epsilon)
\widehat A_h\right)\\
&+\beta\bigl(V_\psi(\widehat{\mathbf z}_h)
-G_h^\lambda\bigr)^2
-\eta\,\mathcal H\bigl(\pi_\theta(\cdot\mid\widehat{\mathbf z}_h)\bigr)
\Big].
\end{aligned}
\end{equation}
Here, $\epsilon$ is the clipping parameter, $\beta$ the value loss coefficient, and $\eta$ the entropy coefficient. 

\dwellcheck{AUTHOR-RUNS}{Populate the final run configuration/data table from
the experiments: initial-data and actor-initialization provenance; transitions
per task; train/validation/test construction; teacher/backbone details; loss
weights and reductions; BPTT and reward/imagination horizons; optimizer rates;
PPO epochs, batch sizes, clipping and entropy/value coefficients; warmup and
refresh budgets. Bind each reported result to its run configuration. These
values and all-task bindings are not supplied by the pasted audits.}

\end{document}

%% file: math_commands.tex
\usepackage{amsmath,amsfonts,bm}

\def\eqref#1{equation~\ref{#1}}
\def\1{\bm{1}}

\DeclareMathAlphabet{\mathsfit}{\encodingdefault}{\sfdefault}{m}{sl}
\SetMathAlphabet{\mathsfit}{bold}{\encodingdefault}{\sfdefault}{bx}{n}